\documentclass[letterpaper, 10 pt, conference]{ieeeconf}  
\IEEEoverridecommandlockouts                              

\usepackage{times}

\usepackage{multicol}
\usepackage[bookmarks=true]{hyperref}

\usepackage{comment}

\usepackage{graphicx,import}

\usepackage{mathtools, amssymb,amsthm}  
\usepackage{paralist}
\usepackage[pdf]{svg}

\graphicspath{{imgs/}}

\newtheorem{assumption}{\bf{Assumption}}

\newtheorem{boldLemma}{\bf{Lemma}}
\newtheorem{theorem}{\bf{Theorem}}

\DeclareMathOperator*{\skewop}{skew}

\newcommand{\ls}{\hspace{0em}}     
     
\newcommand{\rmh}{\mathrm{h}}

\DeclareMathOperator*{\argmin}{argmin}
\title{\LARGE \bf
Momentum Control of an Underactuated Flying Humanoid Robot
}

\author{Daniele Pucci, Silvio Traversaro, Francesco Nori
\thanks{*This work was supported by FP7 EU project CoDyCo under Grant 600716 ICT 2011.2.1 Cognitive Systems and Robotics.}
\thanks{All authors belong to the iCubFacility Department, Istituto Italiano di Tecnologia, Genoa 163, Italy 
        {\tt\small firstname.surname@iit.it}}%
}

\begin{document}

\maketitle
\thispagestyle{empty}
\pagestyle{empty}

\begin{abstract}
The paper takes the first step towards the development of a control framework for underactuated flying humanoid robots. These robots may thus have the capacities of \emph{flight}, \emph{contact locomotion}, and \emph{manipulation}, and benefit from technologies  and methods developed  for  \emph{Whole-Body Control} and \emph{Aerial Manipulation}.  As in the case of quadrotors, we assume that the humanoid robot is powered by four thrust forces. For convenience, these forces are  placed at the robot 
hands and feet. The control objective is defined as the asymptotic stabilization of the robot centroidal momentum. This objective allows us  to track a desired trajectory for the robot center of mass  and keep \emph{small} errors between a reference orientation and the robot base frame. Stability and convergence of the robot momentum are shown to be in the sense of Lyapunov. Simulations carried out on a model of the humanoid robot iCub verify the soundness of the proposed approach.


\end{abstract}

\section{Introduction}

The general purpose of providing humanoid robots with some degree of locomotion and manipulation has driven most of the recent research in the humanoid robotics community. Legged and wheeled locomotion, for instance, have proven to be feasible on various humanoid platforms, which can now be envisioned as interfaces for user assistance in several domains~(see, e.g., \cite{handle2017,Wieber2016}). 
The  robot underactuation combined with the (usually) large number of the robot degrees-of-freedom are among the main difficulties for 
humanoid robot locomotion and manipulation control.  
This  paper takes the first step towards the development of platforms having the capacities  of \emph{flight}, \emph{contact locomotion}, and \emph{manipulation}, 
and proposes 
a  framework for underactuated flying humanoid robots. 
Hence, the proposed platforms
belong to the domain of interests of both  \emph{Whole-Body Loco-Manipulation} and \emph{Aerial Manipulation}~\cite{7759225,kondak2015unmanned}, the union of which may be referred to as \emph{Whole-Body Aerial Loco-Manipulation}.



Humanoid and flying robot control has developed along different directions, and suffer from specific limitations. 

Humanoid robot control is often adddressed assuming the robot attached to ground, and in this case the robot is referred to as ~\emph{fixed-base}~\cite{Featherstone2007}. 
To weaken this assumption, one can apply
the Euler-Poincar\`{e} formalism that provides one with singularity free equations of motion for the humanoid robot~\cite[Chapter 13]{marsden2013introduction}. In this case, the robot is referred to as \emph{floating base}. 
When considering these robot equations of motion, however, the mechanical system representing the humanoid robot is underactuated, which forbids  full feedback linearization of the underlying system 
\cite{Acosta05}. The system underactuation is usually dealt with by means of
 constraints arising from the contacts between the robot and the environment. 
 From the control design perspective, instead, a common  strategy for humanoid robots is based on the so called
\emph{stack-of-task}~\cite{Stephens2010}.
These strategies  usually consider
several control objectives organized in a hierarchical or weighted prioritization.
Often, the high-priority  task is the stabilization of the robot momentum~\cite{Stephens2010,koolen2015design,Pucci2016Hum}, whose 
essence  is that of controlling the robot center-of-mass and angular momentum while guaranteeing stable zero-dynamics~\cite{nava2016}. 
Quadratic programming (QP) solvers can be used to monitor  contact forces while achiving  momentum control \cite{Ott2011,Hopkins2015a}. 



Besides classical flight control~\cite{2003_STEVENS}, 
\emph{small and versatile} 
aircraft 
-- referred to as Vertical Take Off and Landing (VTOL) vehicles -- have attracted  the nonlinear control community effort of the last decade (see, e.g.,~\cite{2017_NALDI,2013_HUA}). One of the main assumptions of these work is that the flying robot is powered by a body-fixed thrust force, and moves at \emph{relatively} small  velocities. This latter assumption renders the aerodynamic forces negligible when compared to gravity, and drag effects are but seldom taken into account~\cite{2013_HUA}. 
Then, a common approach for VTOL position control  is the so-called~\emph{vectored-thrust control paradigm}: the aircraft angular velocity is considered as a control input, and its main role is to align thrust and gravity forces.

We believe that there is  a strong technological benefit in bringing humanoid and flying robots closer: a platform combining these two  robot natures may have, at least theoretically, the capacities of \emph{flight}, \emph{contact locomotion}, and \emph{manipulation}, eventually not all used at the same time.  
A first step in this direction has already been taken by the so-called \emph{Aerial Manipulation},
which conceives platforms capable to  \emph{fly} and \emph{manipulate}~\cite{kondak2015unmanned}. Most robots having these two capacities are composed of a VTOL equipped with one or several robotic arms~\cite{6907674,7353534}. These robots, however, are not energetically efficient when flying in confined spaces, and an additional capacity of \emph{contact locomotion} may lower energy consumption considerably. Also, the robot capacity of making contacts with the environment may robustify the achievement of manipulation tasks: an additional end-effector may be used to make a contact with the environment, thus robustifying the overall system against wind gusts. 

This paper takes a further step in bringing humanoid and flying robots closer, and proposes a control architecture for flying humanoid robots. 
We assume that a 
humanoid robot is powered by four thrust forces installed at the robot end effectors, namely the robot hands and feet. Four turbo engines exemplify the robot actuation assumed in this paper. The control objective is the asymptotic stabilization of the robot \emph{centroidal momentum}~\cite{orin}, which allows us to stabilize a reference trajectory for the center of mass and keep \emph{small}, \emph{bounded} errors between a desired orientation and the robot base frame. Reminiscent of the~\emph{vectored-thrust control paradigm} in VTOL control, we assume that the robot joint velocities can be considered as control input, and we then use them to align the effect of thrust forces with gravity effects. Aerodynamic effects are here neglected, and stability and convergence is shown to be in the sense of Lyapunov.  A quadratic programming (QP) solver is employed to consider actuation limits. Finally,  the robot joint torques that stabilize the (desired) joint velocities (in turn achieving momentum control) are generated via an high-gain control at the joint level. Simulation results performed on the humanoid robot iCub verify the soundness of the proposed approach.

The paper is structured as follows. Section~\ref{sec:background} introduces notation, equations of motion, and  robot actuation. Section~\ref{sec:control} describes the strategy for momentum, center of mass position, and orientation control. Simulations verifying the soundness and robustness of the control laws are presented in Section~\ref{sec:simulations}. Remarks and perspectives conclude the paper.

\section{Background}
\label{sec:background}

\begin{figure}[tb]
\centering
\vspace*{2mm}
\def\svgwidth{.7\linewidth}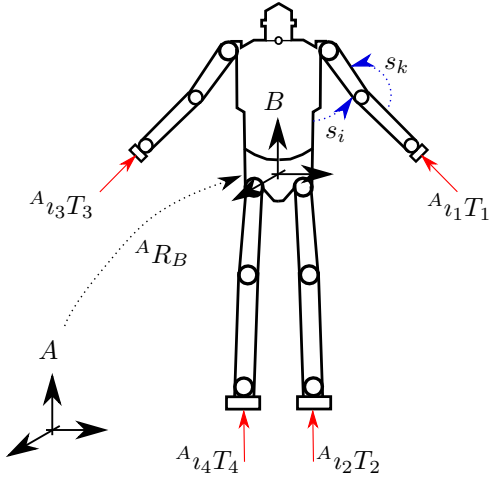
\caption{Notation representation: inertial frame $A$, robot base frame $B$, thrust forces $\prescript{A}{}\imath_k T_k$, $k=\{1,2,3,4\}$, rotation matrix $^AR_B$, and $s_i,s_k$ two of the joint angles, with $i,k \in \mathbb{N}$.}
\label{fig:notation}
 \vspace{-0.05cm} 
\end{figure}

\subsection{Notation}
\begin{itemize}
   \item $A$ denotes an inertial (absolute) frame, composed of a point (\emph{origin}) and an \emph{orientation frame} with its $z$ axis pointing against the gravity, and $g\in \mathbb{R}$  the norm of gravitational acceleration. Given two orientation frames $A$ and $B$, and vectors of coordinates expressed in these orientation frames, i.e. $\prescript{A}{}o,\prescript{B}{}p \in \mathbb{R}^3$, respectively, the rotation matrix 
    $\prescript{A}{}R_B$ is such that $\ls^A p = \ls^A R_B  \ls^B p + \ls^A o_B $. 
    \item $1_n \in \mathbb{R}^{n \times n}$ is the identity matrix of size $n$; $0_{m \times n} \in \mathbb{R}^{m \times n}$ is the zero matrix of size $m \times n$ and $0_{n } = 0_{n \times 1}$.
    \item  $S(x) {\in} \mathbb{R}^{3 \times 3}$ is the skew-symmetric matrix such that $S(x)y {=} x {\times} y$, with $\times$  the cross product operator in $\mathbb{R}^3$. 
\end{itemize}

\subsection{Robot modelling}
We assume that the humanoid robot is composed of $n{+}1$ rigid bodies -- called links -- connected by $n$ joints with one degree of freedom each. 
The 
configuration space of the multi-body system can then be characterized by the \emph{position} and the \emph{orientation} of a frame attached to a robot link -- called 
\emph{base frame} $B$~-- and the joint configurations. More precisely, the robot configuration space  is defined by:
    $\mathbb{Q} = \mathbb{R}^3 \times SO(3) \times \mathbb{R}^n.$
An element of the set $\mathbb{Q}$ is  a triplet $q = (\prescript{A}{}o_{B},\prescript{A}{}R_{B},s)$, where $(\prescript{A}{}o_{B},\prescript{A}{}R_{B})$ denotes the origin  and orientation of the \emph{base frame} expressed in the inertial frame, and $s$ --~which characterizes the robot \emph{shape}~--
denotes the \emph{joint angles}. 
The 
\emph{velocity} of the multi-body system can then be characterized by the set $\mathbb{V}$ defined by:
    $\mathbb{V} = \mathbb{R}^3 \times \mathbb{R}^3 \times \mathbb{R}^n$.
An element of $\mathbb{V}$ is then 
 $\nu = (\mathrm{v}_{B},\dot{s})$, where  $\mathrm{v}_{B} = (^A\dot{ o}_{B},^A\omega_{B})$ is the linear and angular velocity of the base frame  w.r.t. the frame $A$, i.e. $^A\dot{R}_{B} = S(^A\omega_{B})^A{R}_{B}$. 


Applying  the Euler-Poincar\'e formalism \cite[Ch. 13.5]{Marsden2010} to the robot  yields the following equations of motion: 
\vspace{-0.1cm}
\begin{IEEEeqnarray}{RCL}
    \label{eq:system_dynamics}
       {M}(q)\dot{{\nu}} + {C}(q, {\nu}) {\nu} + {G}(q) =  
	\begin{bmatrix} 0_{6} \\ \tau \end{bmatrix}        
         + \sum_{k = 1}^{m} {J}^\top_{k} F_k,
\vspace{-0.1cm}
\end{IEEEeqnarray}
where ${M}, {C}\in \mathbb{R}^{n+6 \times n+6}$ are the mass and Coriolis matrix, respectively,  ${G} \in \mathbb{R}^{n+6}$ is the gravity 
vector, 
$\tau$ are the internal actuation torques, and $F_k \in \mathbb{R}^3$ is the $k$th of the $m$ 
external forces applied by the environment on robot. In particular, we assume that the application point of the external force $F_k$ is the 
origin of a frame $\mathcal{C}_k$, which is attached to the robot's link where the force acts;
the external force $F_k$ is expressed in a frame whose orientation coincides with that of the inertial frame $A$.
The Jacobian ${J}_k= {J}_k(q)$ is the map between the robot's velocity ${\nu}$ and the linear 
velocity 
$ ^A\dot{ o}_{\mathcal{C}_k} \in \mathbb{R}^3$ 
of the origin of 
$\mathcal{C}_k$, i.e.
$^A\dot{ o}_{\mathcal{C}_k} = {J}_{k}(q) {\nu}.$
\subsection{Robot actuation}
\label{subsec:actuation}
We assume that the robot is powered by four thrust forces $T_1,T_2,T_3,T_4\in \mathbb{R}$ that act along the directions 
$\prescript{A}{}\imath_1,\prescript{A}{}\imath_2,\prescript{A}{}\imath_3,\prescript{A}{}\imath_4 \in \mathbb{R}^3$, with 
$|\prescript{A}{}\imath_i| =1$, $\forall i \in \{1,2,3,4\}$, respectively. The application points $\prescript{A}{}o_i\in \mathbb{R}^3$ of the thrust forces are the origins of four 
frames attached to the robot end-effectors, e.g., the robot hands and foot. The  thrust force directions, instead, move accordingly to the robot end-effectors, since thrust forces are assumed to be attached to the end-effector links. Figure~\ref{fig:notation} depicts the notation used for the robot actuation.
Four turbo engines installed at the robot end-effectors exemplify the actuation assumed for the humanoid robot\footnote{
For the humanoid robot iCub, we are considering the turbo engine JETCAT P100 RX~\cite{jetcatP100Rx} 
(weight: 1050gr; length: 245mm; thrust: 100N @152000 1/min).
Note that the weight of  iCub is about 30 Kg, which means that four engines suffice for hovering when fuel is kept off board. 
}. We also assume that each thrust force $T_i$ is measurable. This latter assumption holds, for instance, when force sensors are installed in series with the turbo engines. 

By defining $T:=(T_1,T_2,T_3,T_4)$, the effects of the external thrust forces on the right hand side of the equations of motion~\eqref{eq:system_dynamics} can  be compactly written as follows:
\begin{IEEEeqnarray}{RCL}
    \label{eq:extForce}
       \sum {J}^\top_{k} F_k =  \sum {J}^\top_{k}(q) \prescript{A}{}\imath_k(q)T_k := f(q,T).
\end{IEEEeqnarray}

\section{Control Design}
\label{sec:control}

\subsection{Problem statement}
In light of section~\ref{subsec:actuation}, the  equations of motion 
\begin{IEEEeqnarray}{RCL}
    \label{eq:system_dynamicsF}
       {M}(q)\dot{{\nu}} + {C}(q, {\nu}) {\nu} + {G}(q) &=& 
	\begin{bmatrix} 0_{6 } \\ \tau \end{bmatrix}        
        + f(q,T),
\end{IEEEeqnarray}
are powered by $n+4$ control inputs. This in turn implies that controlling the entire robot configuration space 
$\mathbb{Q} = \mathbb{R}^3 \times SO(3) \times \mathbb{R}^n$, which is of \emph{dimension}\footnote{The set $\mathbb{Q}$ is a Lie Group. Hence, the dimension of $\mathbb{Q}$ is the dimension of the algebra associated with it.} $n+6$, may not be straightforward being  system~\eqref{eq:system_dynamicsF} underactuated and thus not feedback linearisable. Let us remark that  one may attempt at the stabilization of the configuration space $\mathbb{Q}$ by applying advanced techniques developed for underactuated systems evolving on Lie Groups~\cite{ms03,ms04-siam}. The application of these techniques, however, is beyond the scope of this paper.

Assume that the control objective is the asymptotic stabilization of a frame --~i.e. origin and orientation ~-- associated with a robot link. Without loss of generality, assume that one wants to control the base frame of the humanoid robot. The equations of motion of the base frame are given by:
\begin{IEEEeqnarray}{RCL}
   \label{eq:baseFrameDyanmics}
     \dot{\mathrm{v}}_{B} &=& \dot{J}_B \nu + J_B \dot{\nu},
\end{IEEEeqnarray}
with ${\mathrm{v}}_B \in \mathbb{R}^6$ the linear and angular velocity of the base frame, $J_B \in \mathbb{R}^{6\times n+6}$ the Jacobian of the base frame, and $\dot{\nu}$ the robot acceleration obtained from the robot equations of motion~\eqref{eq:system_dynamicsF}. More precisely, by using  the equations~\eqref{eq:system_dynamicsF}, one can evaluate the robot acceleration $\dot{\nu}$ and substitute it into Eq.~\eqref{eq:baseFrameDyanmics}, thus obtaining an instantaneous relationship between the base acceleration $\dot{\mathrm{v}}_{B}$ and the control inputs $(\tau,T)$, i.e.
\[\dot{\mathrm{v}}_{B} = \mathrm{v}(q,\nu,T,\tau).\]
Then, one may attempt at the control of a reference frame by performing feedback linearisation of the above function with  proper feedback correction terms to achieve asymptotic stability. One of the main drawbacks of this approach is that the joint torques $\tau$ have \emph{little} influence on the base acceleration $\dot{\mathrm{v}}_{B}$, which may render the associated control laws ill-posed especially close to constant reference position-and-orientation for the base frame. As a matter of fact, at \emph{low} joint velocities, a first approximation for the  robot equations of motion is given by the Newton-Euler equations, which corresponds to assuming that the robot behaves as a single rigid body. In this case, the dynamics $\dot{\mathrm{v}}_{B}$ can  be obtained from  the Newton-Euler equations, which clearly do not depend upon the (internal) joint torques $\tau$ but only on the four (external) thrust forces. Having only four  inputs  in this dynamics, one cannot perform feedback linearisation of the six-dimensional dynamics $\dot{\mathrm{v}}_{B}$.

Now, since the mass matrix $M(q)$ is positive definite, and thus invertible, the equations~\eqref{eq:system_dynamicsF} point out that the joint dynamics $\ddot{s}$ can be feedback-linearized via a proper choice of the joint torques $\tau$.  So, any differentiable, desired  joint velocity $\dot{s}_d\in \mathbb{R}^n$ can be stabilized with any desired settling time.
We then make the following assumption.

\begin{assumption}
\label{hp:actuation}
The joint velocity $\dot{s}:=u_2$ and  the thrust forces rate-of-change, i.e. $\dot{T}:=(\dot{T}_1,\dot{T}_2,\dot{T}_3,\dot{T}_4):=u_1$, can be chosen at will and then considered as control inputs.
\end{assumption}

In the language of Automatic Control, 
assuming $\dot{s}$ as control variable  is a typical \emph{backstepping} assumption. Then, the production of  the joint torques associated with the desired joint velocities 
can be achieved via classical nonlinear control techniques~\cite[p. 589]{kh02} or high-gain control. 

\subsection{Control objective and centroidal momentum dynamics}
This section shows that despite the aforementioned underactuation, it is possible to conceive control laws for the Newton-Euler equation of the multibody system without any approximation. Then, the control objective for the reminder of this section is  defined as follows:
\begin{itemize}
 \item Asymptotic stabilization of the robot centroidal momentum $\rmh \in \mathbb{R}^6$ about desired, smooth values $\rmh_d(t)\in \mathbb{R}^6$.
\end{itemize}
When complemented with integral correction terms, the laws for the momentum control allows us to: 

$i)$ stabilize a reference trajectory $r(t) \in \mathbb{R}^3$ for the robot center of mass $^A c \in \mathbb{R}^3$; 

$ii)$  keep \emph{small, bounded} errors between a
 reference orientation $R_d(t) \in SO(3)$ and the robot base frame.

Note that the choice of the above objective renders the dimension of the control task equal to six. This means that under Assumption~\ref{hp:actuation}, we may attempt at achieving this task by means of the control inputs $(\dot{s},\dot{T})$. 
More precisely, let $\tilde{\rmh}\in \mathbb{R}^3$ denote the momentum error defined by 
\begin{IEEEeqnarray}{RCL}
    \label{eq:momentumError}
       \tilde{\rmh} := \rmh - \rmh_d(t).
\end{IEEEeqnarray}
Then, by recalling that the rate-of-change of the centroidal momentum equals the summation of all external wrenches acting on the robot~\cite{Orin2013}, one has the following  dynamics:
\begin{IEEEeqnarray}{RCL}
    \label{eq:momentumErrorDynamics}
       \dot{\tilde{\rmh}} &=& A(q)T - mg e_3 - \dot{\rmh}_d(t), \IEEEeqnarraynumspace \IEEEyessubnumber \label{eq:momentumDyn} \\ 
       \dot{T} &=& u_1  \IEEEeqnarraynumspace  \quad \dot{s} = u_2  \IEEEeqnarraynumspace \IEEEyessubnumber \label{eq:inputs} 
\end{IEEEeqnarray}
with $e_3 := (0,0,1,0,0,0)^\top$,
\begin{IEEEeqnarray}{RCL}
    \label{eq:momentumErrorDynamicsComponents}
        A(q) &{=}& 
        \begin{pmatrix}
        \bar{S}(r_{1})\prescript{A}{}\imath_1, \ \bar{S}(r_{2})\prescript{A}{}\imath_2, \ \bar{S}(r_{3})\prescript{A}{}\imath_3, \ \bar{S}(r_{4})\prescript{A}{}\imath_4 
        \end{pmatrix} \IEEEeqnarraynumspace \IEEEyessubnumber \\ 
        r_{i} &{:=}& \prescript{A}{}o_i - \prescript{A}{}c, \quad  \forall i \in \{1,2,3,4\} \IEEEeqnarraynumspace \IEEEyessubnumber \\ 
         \bar{S}(r_{i}) &{:=}& 
        \begin{pmatrix}
        1_3 \\
        S(r_{i})
        \end{pmatrix}.  \IEEEeqnarraynumspace \IEEEyessubnumber 
\end{IEEEeqnarray}
Eq.~\eqref{eq:momentumDyn} points out that the dynamics $\dot{\tilde{\rmh}} \in \mathbb{R}^6$ are underactuated even if one assumes that the thrust intensities $T\in \mathbb{R}^4$ can be considered as control input, and this renders the control of these dynamics  not straightforward. 

The equation~\eqref{eq:momentumDyn}  highlights also that  at the equilibrium configuration $(\tilde{\rmh},\dot{\tilde{\rmh}}){=} (0,0)$, the effect of the thrust intensities must oppose the gravity plus $\dot{\rmh}_d(t)$, i.e.
\begin{IEEEeqnarray}{RCL}
    \label{eq:thrustDirParad}
       0 &=& A(q)T - mge_3 - \dot{\rmh}_d(t). \IEEEeqnarraynumspace  \label{eq:thrustPara} 
\end{IEEEeqnarray}
The above equation is reminiscent of the so-called \emph{vectored-thrust control paradigm} used in recent flight dynamics control techiniques. In this literature, in fact, the aircraft angular velocity is assumed as control input, and then exploited to align the thrust force against the effect of gravity and, in general, of external forces~\cite{2013_HUA,2017_NALDI}. This in turn emphasizes  the role of the joint velocities $\dot{s}$ in establishing Eq.~\eqref{eq:thrustPara}: they are in charge of aligning the total thrust force $A(q)T$ to the gravity and desired momentum rate-of-change effect. 

More precisely and generally, define
\begin{IEEEeqnarray}{RCL}
    \label{eq:definitions}
       \tilde{\xi} &:=& A(q)T + F \IEEEeqnarraynumspace \IEEEyessubnumber \label{eq:xiTilde} \\ 
       F &:=& -mge_3 - \dot{\rmh}_d(t) + K_D \tilde{\rmh} + K_P I(t),  \IEEEyessubnumber   \IEEEeqnarraynumspace \label{eq:F}
\end{IEEEeqnarray}
with $K_P,K_D {\in} \mathbb{R}^{6\times 6}$ two symmetric positive definite matrices, and the variable $I(t)$ --representing the integral of $\tilde{\rmh}$-- governed by $\dot{I}{=} \tilde{\rmh}$. Then,  Eqs.~\eqref{eq:momentumErrorDynamics} can be rewritten as follows:
\begin{IEEEeqnarray}{RCL}
    \label{eq:momentumErrorDynamicsNew}
    	\dot{I} &=& \tilde{\rmh} \IEEEeqnarraynumspace \IEEEyessubnumber \label{eq:Idot} \\
       \dot{\tilde{\rmh}} &=& \tilde{\xi} - K_D \tilde{\rmh} - K_P I, \IEEEeqnarraynumspace \IEEEyessubnumber \label{eq:htildeDot} \\ 
       \dot{\tilde{\xi}} &=& Au_1+\Lambda_s u_2+\Lambda_b \mathrm{v}_{B} 
         - \ddot{h}_d(t) + K_D \dot{\tilde{\rmh}} + K_P \tilde{\rmh}  \IEEEeqnarraynumspace \IEEEyessubnumber \label{eq:xiTildeDot} 
\end{IEEEeqnarray}
with the matrix $A$ given by~\eqref{eq:momentumErrorDynamicsComponents}, 
\begin{IEEEeqnarray}{RCL}
    \label{eq:Lambdabs}
      \Lambda_b &:=& \Lambda 
	  \begin{pmatrix}
	  1_6 \\
	  0_{n\times 6}
	  \end{pmatrix},	        \quad
      \Lambda_s := \Lambda 
	  \begin{pmatrix}
	  0_{6\times n} \\
	  1_{n}
	  \end{pmatrix},	        
         \IEEEeqnarraynumspace \IEEEyessubnumber \label{eq:Lambdab} \\
      \Lambda &:=& -
	  \begin{pmatrix}
	  \tilde{S}_1,\tilde{S}_2,\tilde{S}_3,\tilde{S}_4
	  \end{pmatrix}J_r,	        
         \IEEEeqnarraynumspace \IEEEyessubnumber \label{eq:Lambda}\\
      \tilde{S}_i &:=& 
	  \begin{pmatrix}
	  0_3 && S(^A\imath_i) \\
	  T_iS(^A\imath_i) && T_iS(r_i)S(^A\imath_i)
	  \end{pmatrix},	        
         \IEEEeqnarraynumspace \IEEEyessubnumber \label{eq:STilde}
\end{IEEEeqnarray}
and $J_r$ the Jacobian matrix mapping the robot velocity $\nu$ into the velocities 
$\Omega:= (\dot{r}_1,\omega_1,\dot{r}_2,\omega_2,\dot{r}_3,\omega_3,\dot{r}_4, \omega_4)\in \mathbb{R}^{24}$, i.e. 
      $\Omega = J_r(q)\nu,$
where $\omega_i \in \mathbb{R}^3$ is the angular velocity associated to the $i$th end-effector frame, i.e. $\dot{\prescript{A}{}\imath_i} = S(\omega_i)\prescript{A}{}\imath_i $.

\subsection{Momentum control}
In view of the dynamics~\eqref{eq:momentumErrorDynamicsNew}, the main role of the control inputs $(u_1,u_2)$ is to bring the variable $\tilde{\xi}$ to zero, which means that the effect of the thrust forces must oppose the \emph{apparent} force $F$ (see Eq.~\eqref{eq:xiTilde}). To introduce the control laws accomplishing this task and stabilizing the desired momentum $h_d(t)$, let us recall that the centroidal momentum $\rmh$ is linear versus the robot velocity $\nu$. Namely, there exists a Jacobian matrix $J_\rmh(q) \in \mathbb{R}^{6 \times n +6}$ such that 
\begin{IEEEeqnarray}{RCL}
    \label{eq:CMM}
      \rmh &=& J_\rmh(q)\nu = 
	  \begin{pmatrix}
	  J^b_\rmh& J^s_\rmh
	  \end{pmatrix} \nu =
      J^b_\rmh(q)\mathrm{v}_{B} + J^s_\rmh(q)u_2, \IEEEeqnarraynumspace
\end{IEEEeqnarray}
with $J^b_\rmh \in \mathbb{R}^{6\times 6}$ an invertible matrix, and $J^s_\rmh \in \mathbb{R}^{6\times n}$. The matrix $J_\rmh$ is usually referred to as \emph{centroidal momentum matrix}. We can now present the control laws for system~\eqref{eq:momentumErrorDynamicsNew}. 

\begin{theorem}
\label{th:stability}
Assume that Assumption~\ref{hp:actuation} holds and define 
\begin{IEEEeqnarray}{RCL}
    \label{eq:delta}
      \delta &{:=}& \left(\Lambda_b {+} \tilde{K} J^b_\rmh \right)\mathrm{v}_{B}+\left(K_D{+}1_3\right) \dot{\tilde{\rmh}}  {+} K_P I -\ddot{\rmh}_d  {-}\tilde{K} {\rmh}_d,  \IEEEeqnarraynumspace \IEEEyessubnumber \\ 
      B &{:=}& \Lambda_s {+} \tilde{K} J^s_\rmh,  \IEEEeqnarraynumspace \IEEEyessubnumber \\
      \tilde{K} &{:=}& K_P + K_D + K^{-1}_O,  \IEEEeqnarraynumspace \IEEEyessubnumber 
\end{IEEEeqnarray}
with $K_O \in \mathbb{R}^{6\times 6} $ a symmetric positive definite matrix. 

\noindent
If  there exist 
smooth  
control inputs $(u_1,u_2) {\in} \mathbb{R}^{4+n}$ such that
\begin{IEEEeqnarray}{RCL}
    \label{eq:stabilityCondition}
      \delta + Au_1 + Bu_2 = 0_6,
\end{IEEEeqnarray}
then
the closed loop equilibrium point $(I,\tilde{\rmh},\tilde{\xi}) = (0,0,0)$ of system~\eqref{eq:momentumErrorDynamicsNew} is 
globally asymptotically stable.
\end{theorem}

The proof is given in the Appendix. The main condition of the above theorem is Eq.~\eqref{eq:stabilityCondition}. Let us remark that the  number of control inputs to satisfy this condition is $n+4$, which means that as long as 
\[ \text{rank}(A \ B) = 6,\]
one is left with a redundancy of dimension $n-2$ to render Eq.~\eqref{eq:stabilityCondition} satisfied. This redundancy is later exploited to attempt  at the stabilization of the system \emph{zero dynamics}, i.e. the evolution of system~\eqref{eq:system_dynamicsF} at $(I,\tilde{\rmh},\tilde{\xi}) = (0,0,0)$. 

The explicit form of the state feedback laws, namely $(u_1,u_2) = C(q,\nu,t)$, is here omitted because the expression~\eqref{eq:stabilityCondition} is later exploited into the formulation of a \emph{quadratic programming problem}, which allows us to take into account inequality constraints on the control inputs $(u_1,u_2)$. 

Let us remark that the control laws satisfying Eq.~\eqref{eq:stabilityCondition} are similar to those obtained by applying pure-feedback linearization techniques with output $I$. In particular, the output $I$ is of \emph{relative degree} equal to three, and one can then apply feedback linearization on the obtained dynamics. The laws  deduced with this approach, however, usually forbid to choose control gains independently from each other when the relative degree is higher than two. Also, simulations we have performed tend to show that the law proposed here are more robust w.r.t. to modeling errors than those obtained from pure feedback linearization. 

\subsection{Velocity and position  control}

The control laws satisfying Eq.~\eqref{eq:stabilityCondition} can also be used to stabilize a (smooth) desired velocity $v_{d}(t) \in \mathbb{R}^3$  for the robot center-of-mass. To this purpose, let us recall that the centroidal momentum $\rmh$ can be decomposed into a linear and angular component $h^l, h^\omega \in \mathbb{R}^3$, respectively. Recall also that the linear component $h^l$ is given the robot center-of-mass velocity $\dot{^A c}$ times its mass $m$. In formule, 
\begin{IEEEeqnarray}{RCL}
    \label{eq:momentumDecomp}
     \rmh &=& 
     \begin{pmatrix}
     h^l \\
     h^\omega
     \end{pmatrix} = 
     \begin{pmatrix}
     m \ls^A \dot{c}\\
     h^\omega
     \end{pmatrix}.
\end{IEEEeqnarray}
Consequently, the control laws stabilizing a desired velocity $v_{d}$ for the robot center-of-mass are those satisfying Eq.~\eqref{eq:stabilityCondition} with the desired momentum $\rmh_d(t)$ defined as follows:
\begin{IEEEeqnarray}{RCL}
    \label{eq:desMomVel}
     {\rmh}_d &=& 
     \begin{pmatrix}
     m v_d \\
     {h}_d^\omega
     \end{pmatrix},
\end{IEEEeqnarray}
with $h_d^\omega \in \mathbb{R}^3$ the desired angular momentum. Clearly,  stability and convergence statements of Theorem~\ref{th:stability} are retained under the same assumptions.

Analogously, one can exploit the control laws satisfying Eq.~\eqref{eq:stabilityCondition} to stabilize a (smooth) desired position $r(t) \in \mathbb{R}^3$ for the robot center-of-mass. In particular, it suffices to choose the desired momentum $\rmh_d$ as in Eq.~\eqref{eq:desMomVel} with $v_r = \dot{r}$ and to set the integral initial condition $I(0)$ such that 
\begin{IEEEeqnarray}{RCL}
    \label{eq:integralPosControl}
     I(t) &=& 
     \begin{pmatrix}
     m(^A c - r(t))\\
     I^\omega(0) + \int_0^\infty{\tilde{h}^\omega(s)}ds
     \end{pmatrix}.
\end{IEEEeqnarray}
Again,  stability and convergence statements of Theorem~\ref{th:stability} are retained under the same assumptions.

\subsection{Orientation  control}
This section proposes modifications to Eq.~\eqref{eq:stabilityCondition}  for the  control of the robot base frame $\prescript{A}{}R_{B} \in SO(3)$ towards desired values $R_d \in SO(3)$. Let us first make a short digression on orientation control on $SO(3)$.

The problem of stabilizing a desired orientation $R_d{\in}SO(3)$ for a rigid body orientation $R$ may not be straightforward. 
For instance, 
it is known that 
the topology of $SO(3)$ forbids the design of smooth controllers that globally asymptotically stabilize a reference orientation~$R_d$~\cite{Bhat200063}. Then, quasi-global asymptotic stability is a common feature that orientation controllers guarantee. Just to recall a result:

\begin{boldLemma}[~\cite{Olfati-Saber2000} p. 173] Let  \emph{skew}$(A){:=}\tfrac{1}{2}(A {-} A^\top)$ for any matrix $A\in \mathbb{R}^3$, and the operator $(\cdot)^{\vee}$  defined by $x = S(x)^{\vee}$. Consider the orientation dynamics 
     $\dot{R} = S(\omega)R,$
where $\omega\in \mathbb{R}^3$ is considered as control input. Assume that the control objective is the asymptotic stabilization of a (constant) desired attitude $R_d \in SO(3)$. Then, 
\begin{IEEEeqnarray}{RCL}
    \label{eq:vControlR}
     \omega = -k \left( \skewop(R^\top_d R) \right)^\vee, \quad k>0,
\end{IEEEeqnarray}
renders the equilibrium point $R = R_d$ quasi globally stable. 
\end{boldLemma}
Let us 
recall that to evaluate the control torques generating the angular velocity~\eqref{eq:vControlR}, the correction terms~\eqref{eq:vControlR}  are first  multiplied by the body inertia, and then complemented with additional velocity correction terms~\cite[p. 178]{Olfati-Saber2000}.

The orientation control for the  robot base frame proposed in this paper builds upon the above digression, and consists in modifying the   term $I$ in~\eqref{eq:stabilityCondition} so as to take into account the orientation correction term~\eqref{eq:vControlR}. More precisely, recall that the robot centroidal angular momentum  $h^\omega$ can be expressed in terms of the total robot inertia $\mathbb{\bar{I}}\in \mathbb{R}^{3\times 3}$ as follows,
     $h^\omega = \mathbb{\bar{I}}(q) \omega_o,$
where $\omega_o \in \mathbb{R}^3$ is the so-called  \emph{locked} angualr velocity~\cite{orin}: the terminology encompasses the fact that when joint velocities are blocked, $\dot{s} \equiv 0$, then $\omega_o$ corresponds to the angular velocity of the humanoid robot, which behaves as a  rigid body. This suggests that the control laws~\eqref{eq:stabilityCondition} with 
\begin{IEEEeqnarray}{RCL}
    \label{eq:integralOriControl}
     I &=& 
     \begin{pmatrix}
      I^l(0) + \int_0^\infty{\tilde{\rmh}^l(s)}ds \\
     \mathbb{\bar{I}}(q) \left( \skewop(R^\top_d \prescript{A}{}R_{B}) \right)^\vee
     \end{pmatrix}
\end{IEEEeqnarray}
can guarantee good tracking performance of the base frame $\prescript{A}{}R_{B}$ towards the reference orientation $R_d$. It is important to emphasize that the asymptotic stability of the equilibrium point containing $\prescript{A}{}R_{B} = R_d$ is not guaranteed in this case. Simulations that we have performed, however, tend to show that the suggested control law for the orientation control of the base frame can guarantee good tracking performance even if asymptotic stability is not guaranteed. Control laws ensuring stability properties of the equilibrium point containing $\prescript{A}{}R_{B} = R_d$  will be the subject of forthcoming studies.

\subsection{Orientation and position control}
In light of the above, the control of the robot center-of-mass and base frame can be attempted by using the control laws satisfying Eq.~\eqref{eq:stabilityCondition} with $\rmh_d$  given by~\eqref{eq:desMomVel}, $v_r = \dot{r}$, and 
\begin{IEEEeqnarray}{RCL}
    \label{eq:integralPosOriControl}
     I &=& 
     \begin{pmatrix}
      m(^A c - r(t))\\
     \mathbb{\bar{I}}(q) \left( \skewop(R^\top_d \prescript{A}{}R_{B}) \right)^\vee
     \end{pmatrix}.
\end{IEEEeqnarray}

\subsection{Zero dynamics and optimization problem}
Now, assume that the humanoid robot center of mass and base frame are stabilized about the reference position and orientation, respectively. This constrains only six out of the $n+6$ \emph{degrees of freedom} of the robot, thus leaving an $n$-dimensional \emph{ free-motion} of the system at the desired values: this free motion should be at least bounded. More precisely, this section discusses how to deal with the \emph{boundedness} of the system zero dynamics by exploiting the input redundancy when satisfying Eq.~\eqref{eq:stabilityCondition}.

Let us recall that stability and convergence in Theorem~\ref{th:stability} are shown when Eq.~\eqref{eq:stabilityCondition} holds, i.e.
      $\delta + Au_1 + Bu_2 {=} 0_6$. 
Finding the control inputs $(u_1,u_2) \in \mathbb{R}^{n+4}$ such that this equation holds in general leaves a $n-2$ dimensional input redundancy, which can be used for other purposes. We here use this redundancy so as the joint velocity $u_2 = \dot{s}$ are \emph{as close as possible} to a \emph{postural task} of the following form:
\begin{IEEEeqnarray}{RCL}
	\label{eq:postural}
     p := -K^p_P(s - s_r),
\end{IEEEeqnarray}
with $K^p_P \in \mathbb{R}^{n\times n}$ a symmetric positive definite matrix, and $s_r \in \mathbb{R}^n$ a reference position for the joint configuration. If 
$u_2 = p$ the joint configurations tend to the reference value $s_r$, thus reducing the risk of  unstable zero dynamics.

We combine the tasks of satisfying Eq.~\eqref{eq:stabilityCondition} and $u_2 = p$ in a weighted optimization problem of the following form:
\begin{IEEEeqnarray}{CCL}
	 \label{eq:optimizationProblem}
      (u^*_1,u^*_2) = \argmin_{(u_1,u_2)} &\lambda_m& | \delta + Au_1 + Bu_2|^2 + \lambda_p | u_2 - p|^2 \IEEEeqnarraynumspace \nonumber \\
                             &+& \lambda_{s} | u_2 |^2 + \lambda_{T} | u_1 |^2 \IEEEyessubnumber \\
                             \text{s.t.}  
                             &lb_1& < u_1 < ub_1, 
                            \quad lb_2 < u_2 < ub_2  \IEEEyessubnumber \IEEEeqnarraynumspace
\end{IEEEeqnarray}
where $\lambda_m, \lambda_p, \lambda_{s}, \lambda_{T}$ are positive weighting constants, and  $lb_1, ub_1 \in \mathbb{R}^{4}$ and $lb_2, ub_2 \in \mathbb{R}^{n}$ are the lower and upper bounds for the thrust-intensity variations $u_1$ and joint velocities $u_2$, respectively. Note that the cost function of the optimization problem~\eqref{eq:optimizationProblem} contains also some regularization terms depending on $| u_2 |^2$ and $ | u_1 |^2$.
 
\subsection{Torque control for joint velocity stabilisation}

The solution to the  problem~\eqref{eq:optimizationProblem} is a pair $(u^*_1,u^*_2)$, namely the instantaneous rate-of-change of the thrust intensities $\dot{T}^*$ and the joint velocities $\dot{s}^*$. This latter value is then interpreted as a desired value to be stabilized by a torque-control law. More precisely, the route we follow is high-gain control for the stabilization of $\dot{s}^*$. 
Now, partition~\eqref{eq:system_dynamicsF} as follows
   $  M~=~\begin{pmatrix}
     \mathbb{I} && \mathbb{F} \\
     \mathbb{F}^\top && \mathbb{H}
     \end{pmatrix}$
with $\mathbb{I} \in \mathbb{R}^{6\times 6}$, $ \mathbb{F} \in \mathbb{R}^{6\times n}$, $\mathbb{H}\in \mathbb{R}^{n \times n}$,
  $   b := 
	\begin{pmatrix}
		b_b \\
		b_s
	\end{pmatrix}	 :=
     C(q,\nu)\nu + G(q) 
     \label{eq:fPartition}
     f :=
	\begin{pmatrix}
		f_b \\
		f_s
	\end{pmatrix}
$ with $b_b,f_b {\in} \mathbb{R}^6$ and $b_s,f_s {\in} \mathbb{R}^n$. Then, from Eq.~\eqref{eq:system_dynamicsF}, one gets
\begin{IEEEeqnarray}{RCL}
     &\bar{M}& \ddot{s} + \bar{b} = \tau   \IEEEyessubnumber  \label{eq:jointDynamics} \\
     \bar{M} &:=& \mathbb{H} -  \mathbb{F}^\top \mathbb{I}^{-1}\mathbb{F}, \quad  
     \bar{b} := b_s - f_s + \mathbb{F}^\top \mathbb{I}^{-1} (f_b-b_b)    \IEEEyessubnumber. \label{eq:jointDynamicsElements} \IEEEeqnarraynumspace
\end{IEEEeqnarray}
In view of Eq.~\eqref{eq:jointDynamics}, the stabilization of a desired joint velocity $u^*_2$ is then attempted by means of the following high-gain torque control law:
\begin{IEEEeqnarray}{RCL}
	 \label{eq:torqueControl}
     \tau =  \bar{b}+\bar{M}\left( K^s_P(u^*_2 - \dot{s}) +  K^s_I \int_0^t(u^*_2 - \dot{s})dt \right).
\end{IEEEeqnarray}

\begin{figure}[t!]
\centering
\includegraphics[width=1\linewidth]{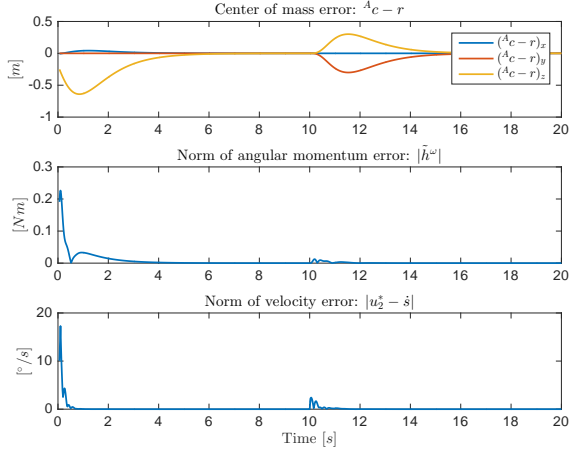}
\caption{Data of simulation 1.
From top to bottom, 
center-of-mass tracking error, norm of angular momentum error, and joint velocity tracking error.}
\label{fig:comErrAngMomVelErrS1}
 \vspace{-1em} 
\end{figure}


\begin{figure}[t!]
\centering
\includegraphics[width=1\linewidth]{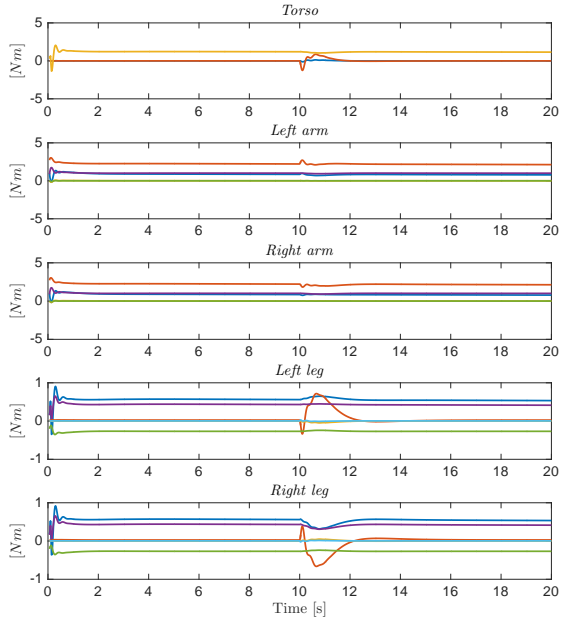}
 \vspace{-2em} 
\caption{Data of simulation 1.
This picture depicts the joint torques $\tau$ necessary for stabilizing the joint velocity error $u^*_2 - \dot{s}$ to zero. From top to bottom, joint torques of the robot torso, left and right arms, and left and right legs.}
\label{fig:jointTorquesS1}
 \vspace{-1em} 
\end{figure}

\section{Simulation Results}
\label{sec:simulations}

This section presents simulations results obtained by applying the control algorithm discussed in the previous section. These simulations have been carried out by using the humanoid robot iCub with 25 DoFs \cite{metta2010icub}.  In particular, the robot joint angles $s \in \mathbb{R}^{25}$ can be partitioned into five robot parts, which correspond to the robot {torso} $s_t \in \mathbb{R}^3$, {left arm} $s_{la} \in \mathbb{R}^5$, {right arm} $s_{ra}\in \mathbb{R}^5$, {left leg} $s_{ll} \in \mathbb{R}^6$, and {right leg} $s_{rl} \in \mathbb{R}^6$, i.e. $s = (s_t,s_{la},s_{ra},s_{ll},s_{rl})$.

\subsection{Simulation environment and control parameters}
\label{sec:num_int_description}
The custom simulation environment is in charge of integrating the equations of motion~\eqref{eq:system_dynamicsF} with the joint torques $\tau$ given by~\eqref{eq:torqueControl} and the thrust intensities $T$ provided by the time integration of $u^*_1$, which are in turn generated by~\eqref{eq:optimizationProblem}.

For time-integration purposes of~\eqref{eq:system_dynamicsF}, we parametrize $SO(3)$ by means of a quaternion representation 
$\mathcal{Q} \in \mathbb{R}^4 $. 
The resulting  state space system, which is integrated through time, is then:
$
\chi :=(
    ^A o_{{B}}, \mathcal{Q}, s, \dot{p}_{{B}}, \omega_{{B}}, \dot{s}
)
$, and its derivative is given by
$
\dot{\chi} = (
    ^A \dot{o}_{{B}},  \dot{\mathcal{Q}}, \dot{s}, \dot{\nu}).
$
The constraints $|\mathcal{Q}|=1$ are  enforced during the integration phase~\cite{Gros2015}.
The system evolution is  obtained by integrating the constrained dynamical system with the numerical integrator MATLAB \emph{ode15s} thanks to our  software abstraction interfaces -- described in~\cite{romano2017} -- which provide us the elements of the equations of motion~\eqref{eq:torqueControl}, e.g. the mass matrix.
\begin{figure}[t!]
\centering
\includegraphics[width=1\linewidth]{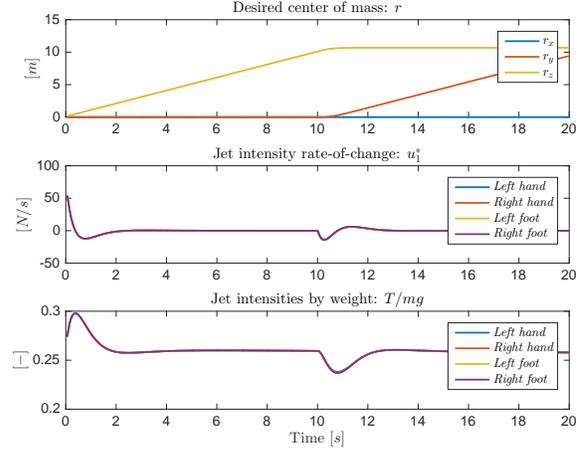}
\caption{Data of simulation 1.
From top to bottom, desired center-of-mass,  jet intensity rate-of-change~$u^*_1$, and  thrust-to-weight ratio obtained by time integration of  $u^*_1$.}
\label{fig:comDesJetsDotJetsS1}
 \vspace{-1em} 
\end{figure}
The humanoid robot initial condition is $\chi(0) = (0_3, 1,0_3, 0_3,0,25,0_3,0,25,0_3,0_{12})$, which corresponds to having the robot arms open ($25^\circ$ on the robot shoulders) and the base frame corresponding to the inertial frame. The robot legs are at the zero position, which corresponds to having the robot legs straight.

We then set the control gains as follows: for the momentum control described in Theorem~\ref{th:stability}, $K_P = (1_3 5, 0_3;0_3,1_350)$, $K_D = 2\sqrt{K_P}$, $K_O = 10$; for the optimization problem~\eqref{eq:optimizationProblem}: $\lambda_m = 50$, $\lambda_p = 1$, $\lambda_s = 50$, $\lambda_T{=}1$, $K^p_P {=}\text{diag}(1, 1, 1, 5, 5, 5, 10, 1, 5, 5, 5, 10, \text{ones}(1,12))$, with limits
$ub_1 = -lb_1 = 45 \ \text{ones}(4,1) \ [^\circ/s]$,  $ub_2 = -lb_2 = 100 \ \text{ones}(25,1) \ [N/s]$;
and for the high-gain torque control~\eqref{eq:torqueControl}: $K^s_I = 10^3 \ \text{diag}(\text{ones}(25,1))$, $K^s_P = 2\sqrt{K^s_I }$.

In the following simulations, noise is not considered. The main reason why is that given the robot actuation, we expect that  the noise produced by turbo engines may be higher than that produced by the robot joint motors. Noise nature and amplitude produced by turbo engines when attached to the humanoid robot is not known for the time being. Hence, 
noise will be considered  after proper experimental campaigns will be carried on to identify
 noise nature and amplitude.


\subsection{Simulation 1: piece-wise constant velocity trajectory without 
orientation control}

The first simulation concerns the stabilization of a reference trajectory $r(t) \ [m/s]$ obtained by integrating the following piece-wise constant reference velocity:

\begin{equation}
\label{eq:refVel}
  \dot{r}(t) =
  \begin{cases}
    (0, \ 0, \ 1) & \text{if }\quad  0 \leq t < 10  \ [s]
    \\
    (1, \ 0, \ 0) & \text{if }\quad  10 \leq t  \quad \quad  \ [s],
  \end{cases}
\end{equation}
with $r(0) = 0_3$. To stabilize the resulting trajectory, we then apply the algorithm~\eqref{eq:optimizationProblem}-\eqref{eq:torqueControl}, with the integral term $I$ given by~\eqref{eq:integralPosControl} and $h^\omega_d = 0_3$. Data associated with this simulation are depicted in Figures~\ref{fig:comErrAngMomVelErrS1}--\ref{fig:comDesJetsDotJetsS1}. In particular, from top to bottom, Figure~\ref{fig:comErrAngMomVelErrS1} depicts  the tracking error of the robot center of mass, the angular momentum error, and  the norm of the joint velocity error $u^*_2-\dot{s}$. Since all tracking errors converge to zero, the simulation results shown in Figure~\ref{fig:comErrAngMomVelErrS1}  verify the statements of Theorem~\ref{th:stability} when combined with the high-gain control~\eqref{eq:torqueControl}. The control gains $K_P$ and $K_O$ influence the settling time with which both tracking errors of the center-of-mass and the angular momentum tend to zero. Still in Figure~\ref{fig:comErrAngMomVelErrS1}, observe also that the norm of the velocity error $u^*_2 - \dot{s}$ tends to zero meaning that the high-gain control~\eqref{eq:torqueControl} for joint velocity stabilization is working properly.  The joint torques~\eqref{eq:torqueControl} to obtain this stabilization  are depicted in Figure~\ref{fig:jointTorquesS1}. Observe that despite the high-gain control chosen in this paper, the joint torques remain relatively \emph{small} during the task $|u^*_2-\dot{s}|\rightarrow 0$.

The outcome $u^*_1$ of the optimization problem~\eqref{eq:optimizationProblem} is depicted in Figures~\ref{fig:comDesJetsDotJetsS1}. Note how the optimizer tends to choose solutions that render equal all thrust intensities variations thanks to the term $\lambda_T$. The smaller this regularization term, the more different the rate-of-change $u^*_1$ of the jet intensities.





\subsection{Simulation 2: helicoidal flight  with orientation control and non-perfect control model to test controller robustness}
\begin{figure}[t!]
\centering
\includegraphics[width=1\linewidth]{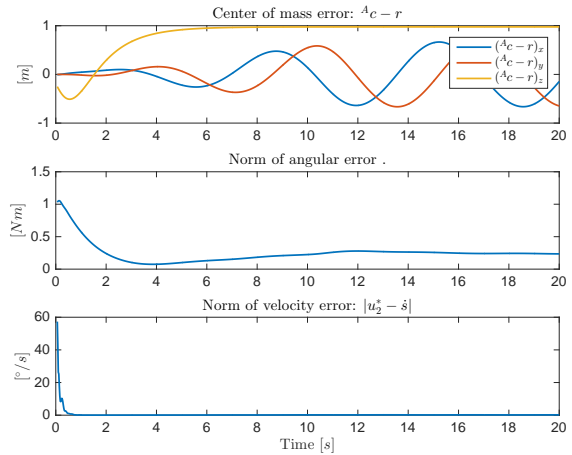}
\caption{Data associated with simulation 2.
From top to bottom, 
center-of-mass tracking error, norm of angular  error, and joint velocity tracking error.}
\label{fig:comErrAngMomVelErr2}
 \vspace{-1em} 
\end{figure}

The second simulation consists in tracking the reference trajectory given by
\begin{IEEEeqnarray}{RCL}
	\label{eq:helicoidal}
	r(t) =  A(t)\cos(0.3\pi t)e_1 +A(t)\sin(0.3\pi t)e_2 + t e_3, \IEEEeqnarraynumspace
\end{IEEEeqnarray}
with 
\begin{equation}
\label{eq:refVelA}
  A(t) = 
  \begin{cases}
    2 \tfrac{t}{10} & \text{if }\quad  0 \leq t < 10  \ [s]
    \\
    2 & \text{if }\quad  10 \leq t  \quad \quad  \ [s]
  \end{cases}.
\end{equation}
After $10 \ [s]$, it  represents a helicoidal trajectory of radius $2 \ [m]$ with vertical speed of $1 \ [m/s]$. Differently to the previous section, we here test the trajectory tracking controller with orientation control. So, we apply~\eqref{eq:optimizationProblem}-\eqref{eq:torqueControl} with the integral term $I$ given by~\eqref{eq:integralPosOriControl} and $R_d$ obtained by rotating $\prescript{A}{}R_B(0)$ about the vertical axis of $60^\circ$. 
Then, the robot is expected to rotate about the vertical axis to reduce the orientation error. 
%
Also, to test the control robustness, the controller has been calculated with estimated parameters that differ from the real ones of $10 \ \%$. More precisely, the controller is evaluated by over-estimating the real model, i.e. $\hat{M}(q) = 1.1M(q)$, $\hat{C}(q,\nu)\nu+\hat{G}(q) = 1.1(C(q,\nu)\nu+G(q))$, and $\hat{J}_h(q) = 1.1 J_h(q)$. As depicted in Figure~\ref{fig:comErrAngMomVelErr2}, modeling errors lead to non convergence of the center-of-mass tracking error  and orientation error to zero. However, \emph{relatively small} bounded errors between actual and reference signals are kept, despite the controller~\eqref{eq:optimizationProblem}-\eqref{eq:torqueControl}~\eqref{eq:integralPosOriControl} is not guaranteed to possess stability and convergence properties. This shows a degree of robustness versus modeling errors of the control laws presented in this paper. Let us remark that the orientation control performance degrades quickly for a time-varying reference orientation, and this calls for studies and extensions of the control laws presented in this paper for time-varying orientation tracking.
Due to lack of space, plots depicting the control inputs are omitted.

\section{Conclusions}
\label{sec:conclusions}
This paper has proposed extensions of the so-called~\emph{vectored-thrust control paradigm} used in VTOL control to the case of an underactuated flying humanoid robot. The main assumption is that the humanoid robot is powered by four thrust forces placed at the robot end effectors: these forces reduce but do not eliminate the well-known humanoid robot underactuation.  Within this actuation framework, we have presented control laws guaranteeing stability and convergence properties for the robot centroidal momentum. Slight modifications to these laws allows us to track a desired reference trajectory for the center of mass and keep \emph{small} tracking errors between a reference orientation and the robot base frame.  

In this respect,  future work consists in proposing control laws guaranteeing stability and convergence not only for the desired center of mass position and angular momentum,  but also for the reference orientation of the base frame. Also, the model of the external forces acting on the robot neglects the aerodynamic phenomena. Hence, future work will also consist in extending~\emph{vectored-thrust control paradigm} with aerodynamic effects to the case of the considered flying humanoid~\cite{2013_PUCCI,pucciPhd,phms15}.
\vspace{-0.1cm}
\section*{Appendix: proof of Theorem~\ref{th:stability}}
\vspace{-0.1cm}

Consider the following Lyapunov function candidate
\begin{IEEEeqnarray}{RCL}
	\label{eq:lyapunovCandidate}
	V(I,\tilde{h},\tilde{\xi}) &:=& \frac{1}{2}I^\top K_P I+\frac{1}{2}|\tilde{\rmh}|^2 + \frac{1}{2}\tilde{\xi}^\top K_O \tilde{\xi}.
\end{IEEEeqnarray} 
Note that $V = 0 \iff (I,\tilde{h},\tilde{\xi}) = 0$. Direct calculations show   $\dot{V}$ along the trajectories of system~\eqref{eq:momentumErrorDynamicsNew} is given by:
\begin{IEEEeqnarray}{RCL}
	\label{eq:lyapunovCandidateDot}
	\dot{V} &=& -\tilde{\rmh} ^\top K_D \tilde{\rmh}  + \tilde{\xi}^\top K_O \Big(\dot{\tilde{\xi}} + K^{-1}_O \tilde{\rmh} \Big) 
\end{IEEEeqnarray} 
Now, it is clear that $\dot{V}\leq 0$  if the term in the parenthesis on the right hand side of the above equation is equal to~$-\tilde{\xi}$. In view of~\eqref{eq:htildeDot} and~\eqref{eq:definitions},  imposing
$ {-} \tilde{\xi}  \nonumber =\dot{\tilde{\xi}} + K^{-1}_O \tilde{\rmh}$ yields:
\begin{IEEEeqnarray}{RCL}
 \label{eq:xiDotClosed} 
	{-} \tilde{\xi} &=&
	Au_1{+}\Lambda_s u_2 {+}\Lambda_b \mathrm{v}_{B} 
         {-} \ddot{\rmh}_d(t) {+} K_D \dot{\tilde{\rmh}} {+} \left( K_P  {+} K^{-1}_O \right) \tilde{\rmh} \nonumber \\
         &=& -A(q)T + mge_3 + \dot{\rmh}_d(t) - K_D \tilde{\rmh} - K_P I \IEEEeqnarraynumspace  	\label{eq:stabilityCondStep1}
\end{IEEEeqnarray} 
By substituting $\tilde{\rmh} = \rmh - \rmh_d$ and
$\rmh~=~J^b_\rmh(q)\mathrm{v}_{B}~+~J^s_\rmh(q)u_2$ in~\eqref{eq:stabilityCondStep1}, one gets the condition~\eqref{eq:stabilityCondition} of Theorem~\ref{th:stability}. As a consequence, if~\eqref{eq:stabilityCondition} -- and consequently~\eqref{eq:stabilityCondStep1} -- are always satisfied, then $\dot{V}$ in Eq.~\eqref{eq:lyapunovCandidateDot} becomes
	$\dot{V} = -\tilde{\rmh} ^\top K_D \tilde{\rmh}  - \tilde{\xi}^\top K_O \tilde{\xi} \leq 0.$
And $\dot{V} \leq 0$ implies the stability of the equilibrium  point and also global boundedness of the system trajectories. 

Now, as long as~\eqref{eq:stabilityCondStep1} is satisfied, the closed loop dynamics is given by $\dot{\tilde{\xi}} = -\tilde{\xi} -K^{-1}_O\tilde{\rmh}$
and~\eqref{eq:Idot}~\eqref{eq:htildeDot}: therefore, the closed loop dynamics is autonomous, and we can use LaSalle Theorem to conclude that  $\dot{V} \rightarrow 0$.  This implies that $\tilde{\rmh} \rightarrow 0$, $\tilde{\xi} \rightarrow 0$, and $\dot{\tilde{\xi}} \rightarrow 0$, $\dot{\tilde{\rmh}} \rightarrow 0$. By using these implications with~\eqref{eq:htildeDot}, namely
     $  \dot{\tilde{\rmh}} = \tilde{\xi} - K_D \tilde{\rmh} - K_P I, $
one obtains that $I \rightarrow 0$. Hence, the equilibrium point $(I,\tilde{\rmh},\tilde{\xi}) = (0,0,0)$ is asymptotically stable. Global asymptotic stability comes from radial unboundedness of $V$. 

\vspace{-0.1cm}
\bibliographystyle{IEEEtran}
\bibliography{references}

\end{document}